\documentclass{article} 
\usepackage{iclr2026_conference,times}

\usepackage{amsmath,amsfonts,bm}

\def\eqref#1{equation~\ref{#1}}

\def\1{\bm{1}}

\DeclareMathAlphabet{\mathsfit}{\encodingdefault}{\sfdefault}{m}{sl}
\SetMathAlphabet{\mathsfit}{bold}{\encodingdefault}{\sfdefault}{bx}{n}

\usepackage[colorlinks, linkcolor=red, anchorcolor=darkblue, citecolor=mygreen2, pagebackref]{hyperref}

\usepackage[nameinlink,noabbrev]{cleveref}

\usepackage{natbib}
\definecolor{DeepPink}{RGB}{255,20,147}
\usepackage{amsmath}
\definecolor{darkblue}{rgb}{0,0.08,0.45}
\definecolor{cvprblue}{rgb}{0.21,0.49,0.74}
\definecolor{mygreen2}{RGB}{0 205 0}
\usepackage{colortbl}
\usepackage{amsfonts,bm,pifont}
\usepackage{marvosym}
\usepackage{placeins}

\usepackage{url}

\usepackage{booktabs}
\usepackage[table]{xcolor}
\usepackage{ulem}
\usepackage{soul, color, xcolor}
\usepackage{graphicx}
\usepackage{wrapfig}
\usepackage{amssymb}
\usepackage{multirow}
\usepackage{array} 
\usepackage{makecell}
\usepackage{algorithm}
\usepackage{algpseudocode}
\usepackage{xspace} 
\usepackage{tabularx}
\usepackage{makecell} 
\usepackage[dvipsnames]{xcolor}
\usepackage{subfig}
\usepackage{wrapfig}
\usepackage{algorithm}
\usepackage{algpseudocode}
\newcolumntype{L}{>{\raggedright\arraybackslash}p{2.6cm}} 
\newcolumntype{C}{>{\centering\arraybackslash}X}

\newcommand{\methodname}{VLA-RFT\xspace}
\title{VLA-RFT: Vision-language-action Reinforcement fine-tuning with verified rewards in world simulators}

\author{
Hengtao Li$^{1,3,7*,\diamondsuit}$ 
Pengxiang Ding$^{1,2,3,*,\dagger}$ 
Runze Suo$^{1,3,4,*}$ 
Yihao Wang$^{1,3,6,*}$
Zirui Ge$^{1,2,3}$ 
\\ 
\textbf{
Dongyuan Zang$^{7}$ 
Kexian Yu$^{5}$ 
Mingyang Sun$^{2,3,4}$ 
Hongyin Zhang$^{1,2}$
Donglin Wang$^{1}$\textsuperscript{\Letter} 
Weihua Su$^{7}$\textsuperscript{\Letter} 
}\\
$^{1}$Westlake University \thinspace 
$^{2}$Zhejiang University \thinspace
$^{3}$OpenHelix Team \thinspace
$^{4}$Fudan University \thinspace \\
$^{5}$Zhengzhou University\quad
$^{6}$BUPT \thinspace
$^{7}$Hebei University of Technology \thinspace
 \\
$^{\dagger}$Project Lead: dingpx2015@gmail.com\\
\textsuperscript{\Letter}Corresponding Author \thinspace 
$^{*}$Equal contribution \thinspace
$^{\diamondsuit}$Work done during interning at Westlake University\\
}

\iclrfinalcopy 
\begin{document}

\maketitle

\begin{figure*}[ht!]
    \centering
    \includegraphics[width=\linewidth]{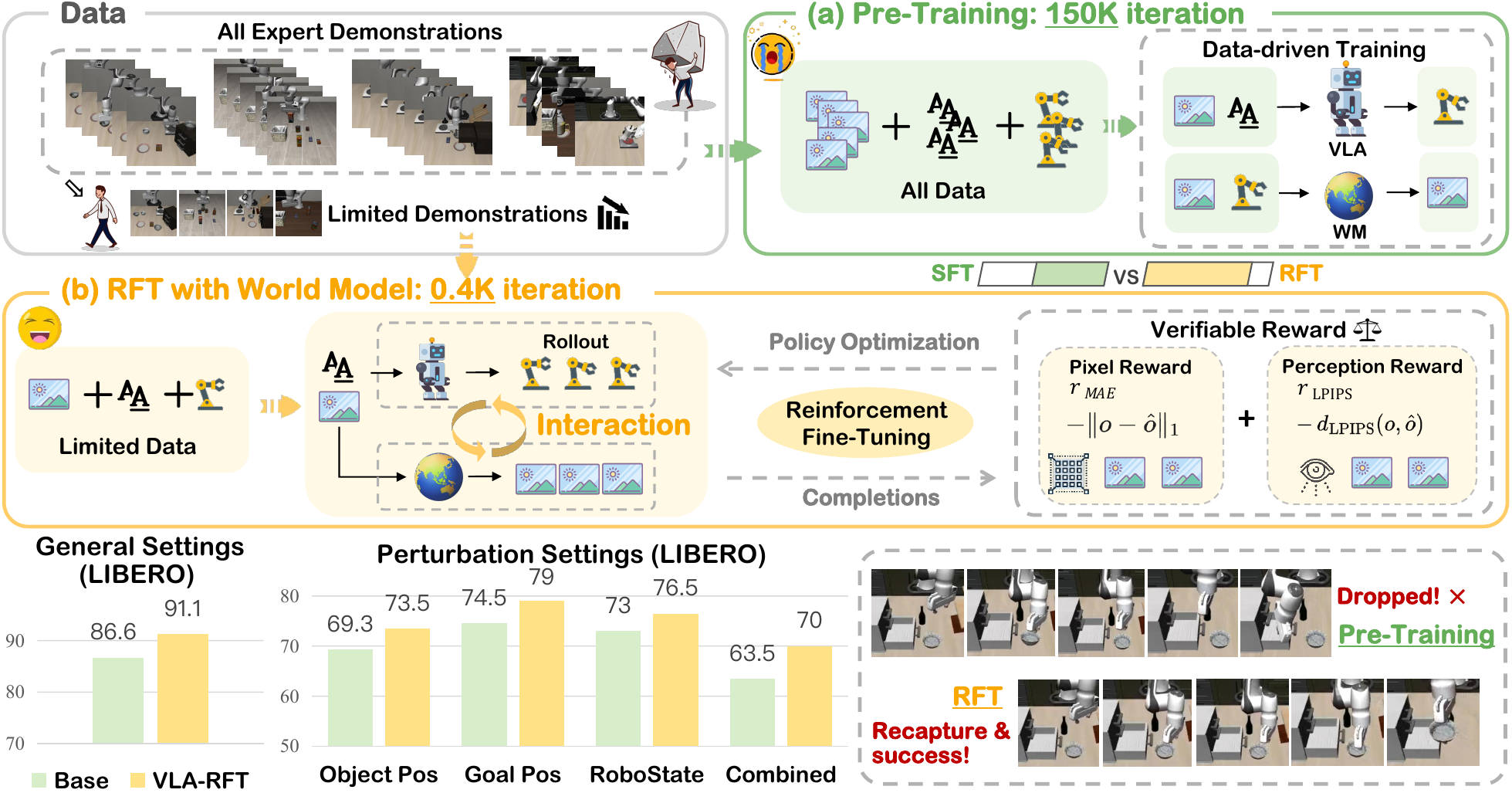}
    \caption{\textbf{The Framework of VLA-RFT.} 
    A world model functions as a simulator that processes multi-rollout VLA action sequences to generate corresponding future states. By incorporating verified rewards through a GRPO optimization framework, we perform end-to-end updates of the VLA. Our approach achieves superior performance with remarkably fewer optimization steps—requiring only 0.4K iterations compared to 150K iterations for a strongly supervised baseline—demonstrating advantages in both standard and perturbed environments. Furthermore, the method exhibits enhanced execution-time robustness, characterized by reliable failure recovery and retry capabilities. For more details, please refer to our \href{https://vla-rft.github.io/}{webpage}.}
    \label{fig:teaser}
\end{figure*}

\begin{abstract}
Vision-Language-Action (VLA) models enable embodied decision-making but rely heavily on imitation learning, leading to compounding errors and poor robustness under distribution shift. Reinforcement learning (RL) can mitigate these issues yet typically demands costly real-world interactions or suffers from sim-to-real gaps. We introduce VLA-RFT, a reinforcement fine-tuning framework that leverages a data-driven world model as a controllable simulator. Trained from real interaction data, the simulator predicts future visual observations conditioned on actions, allowing policy rollouts with dense, trajectory-level rewards derived from goal-achieving references. This design delivers an efficient and action-aligned learning signal, drastically lowering sample requirements. With fewer than 400 fine-tuning steps, VLA-RFT surpasses strong supervised baselines and achieves greater efficiency than simulator-based RL. Moreover, it exhibits strong robustness under perturbed conditions, sustaining stable task execution. Our results establish world-model-based RFT as a practical post-training paradigm to enhance the generalization and robustness of VLA models. For more details, please refer to our \href{https://vla-rft.github.io/}{webpage}.
\end{abstract}
\section{Introduction}

Vision-Language-Action (VLA) models have recently achieved remarkable progress by building upon large, pre-trained vision-language models (VLMs) \citep{eagle, prismatic, palm}. Leveraging the powerful perceptual generalization of VLMs allows these models to operate under diverse visual conditions. However, most existing VLAs \citep{rt1, rt2, pi0, bjorck2025gr00t, openvla} are trained purely via imitation learning. This approach is prone to error accumulation under distribution shift, where small deviations from expert demonstrations gradually drive the policy toward unfamiliar states and weaken its robustness \citep{ross2010efficient, de2019causal, foster2024behavior}.

In contrast, reinforcement learning (RL) offers a promising avenue to overcome these limitations by explicitly optimizing beyond demonstrated behaviors and encouraging exploration \citep{liu2025can}. Recent studies have increasingly incorporated RL into VLA training, demonstrating its critical role in enhancing generalization and long-horizon task performance through offline RL approaches \citep{reinbot, zhang2024grape}, direct real-world RL \citep{xu2024rldg,guo2025irevla}, and simulation-based RL \citep{lu2025vla,tan2025interactive,liu2025can}.

Yet, standard RL pipelines for VLA face steep challenges. Simulation-based RL \citep{chen2025rlrc, chen2025tgrpo, shu2025rftf} often requires millions of interactions and suffers from a pronounced sim-to-real gap. Real-world training \citep{xu2024rldg, mark2024parl, guo2025irevla, chen2025conrft}, on the other hand, is prohibitively costly and can raise safety concerns. Offline RL also remains limited: without interaction with the environment, models are vulnerable to distribution shift and cannot learn from the consequences of their own actions \citep{tan2025interactive}.

To address these challenges, we propose VLA-RFT, a reinforcement fine-tuning framework that leverages a world model as a high-fidelity simulator for policy optimization. At its core, VLA-RFT employs a controllable world simulator that, once trained on a dataset of robot interactions, can predict future visual observations conditioned on an action sequence. Unlike conventional simulation environments restricted to handcrafted scenarios, this simulator is entirely data-driven, capturing the diversity of real-world interactions while avoiding the prohibitive cost and safety risks of training directly in the physical world. For a given task, policy-proposed actions are rolled out within this simulator to generate predicted visual trajectories. These synthetic trajectories then enable the design of a dense, task-grounded reward by comparing them against the visual trajectory from goal-achieving reference trajectory. These rewards are then used to optimize the policy via Generalized Reinforcement Policy Optimization (GRPO), enabling stable and efficient reinforcement fine-tuning.

This design provides a continuous, action-aligned learning signal that substantially reduces the sample complexity of reinforcement fine-tuning. Empirically, we show that with as few as 400 fine-tuning steps, VLA-RFT not only outperforms strong supervised fine-tuning baselines \citep{vlaadapter} in both overall performance and compositional generalization, but also achieves markedly higher efficiency than simulator-based RL algorithms that demand orders of magnitude more interactions. Furthermore, in perturbed or adversarial scenarios, VLA-RFT exhibits superior action robustness, sustaining stable task execution even under unexpected environmental variations. Taken together, this combination of efficiency, generalization, and robustness underscores the practical advantages of our framework for scalable VLA training.

Finally, we hope that our method, experiments, and analysis will motivate future research to harness world models as a general and efficient post-training paradigm for VLAs, thereby substantially enhancing their practicality and accelerating their real-world deployment.

\section{Related Work}

\noindent
\textbf{Vision-Language-Action Models.}
Vision-Language-Action (VLA) models align visual and linguistic inputs with actions through imitation learning on large-scale datasets \citep{oxe, libero, calvin}. Pre-trained VLMs provide generalization, while supervised fine-tuning adapts them to task-specific action spaces \citep{eagle, prismatic, palm}. Recent studies further improve efficiency with lightweight adapters and post-training techniques \citep{openvla-oft, cui2025openhelix, vlaadapter, fan2025longvla, gong2024carp, ding2024quar, ding2025humanoid}.  
However, imitation learning alone is prone to error accumulation under distribution shifts, where minor deviations from expert data push the policy into unfamiliar states and reduce robustness. 
To address this, recent studies incorporate reinforcement learning to improve VLA performance. Our work also falls into this line of research.

\noindent
\textbf{VLA with Reinforcement Learning.}
Reinforcement learning from human feedback has proven highly effective in language models \citep{sheng2025verl, ouyang2022instructgpt}, inspiring RL fine-tuning for vision–language–action (VLA) systems. However, simulation-based RL \citep{chen2025rlrc, chen2025tgrpo, shu2025rftf} requires vast interactions and suffers from the sim-to-real gap, while real-world training \citep{xu2024rldg, mark2024parl, guo2025irevla, chen2025conrft} is expensive and unsafe. Offline RL also struggles with distribution shift and the inability to learn from its own actions \citep{tan2025interactive}. To overcome these limits, we leverage a world model as a data-driven simulator, enabling practical policy optimization without real-world costs or risks.

\noindent
\textbf{World Models.}
World Models learn environment dynamics for planning and control, either via explicit physics \citep{song2024learning, li2024fld, sancaktar2022curious} or latent predictive representations \citep{planet, dreamer, dreamerv3}. Recent extensions integrate multi-modal inputs and guide RL with high-dimensional predictions \citep{wu2023daydreamer, roboticWorldModel}. Advances in generative modeling \citep{videodiffusionmodel, blattmann2023stablediffusion, liu2024sora} have enabled large-scale video-based World Models \citep{bardes2023vjepa, assran2025vjepa2}, later specialized for robotics \citep{zhou2024dinowm, zhou2024robodreamer}. Emerging works further link these models with instruction-conditioned action generation \citep{hu2024vpp, cen2025worldvla, zhong2025flowvla, zhang2025gevrm}. In this work, our world model not only functions as a dynamics simulator, but also provides verified rewards to fine-tune the VLA, enabling rapid and efficient enhancement of the base model’s performance.

\section{Method}
\begin{figure*}[ht!]
    \centering
    \includegraphics[width=\linewidth]{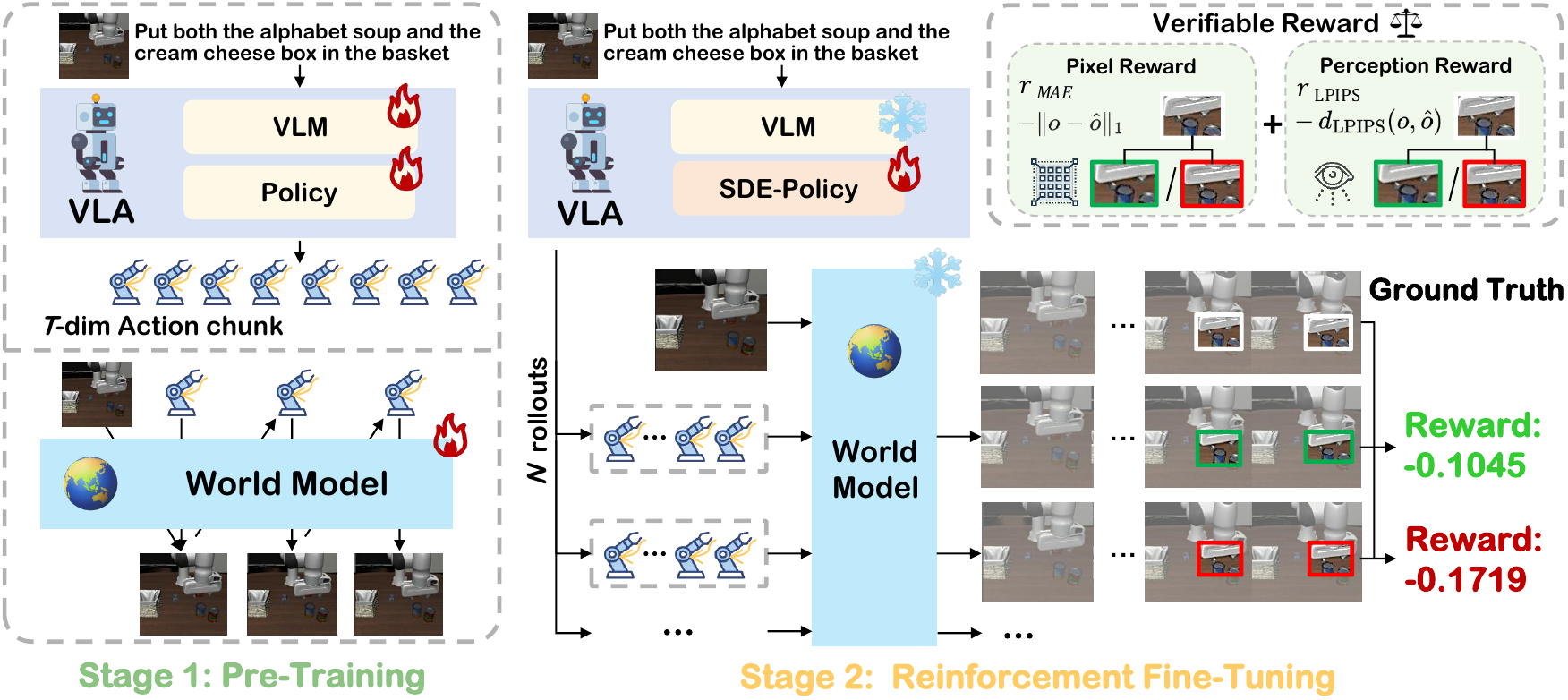}
    \caption{\textbf{Training Paradigm of VLA-RFT.} In the pre-training stage, both the world model and VLA policy are initialized, where the world model takes a 7-dimensional action input that is consistent in format with the VLA’s action output. In the reinforcement fine-tuning stage, the VLA generates action chunks based on an initial frame and language instruction, which are rolled out in the world model to predict future states. Verified rewards are then computed from the predicted states and used to optimize the VLA via GRPO Optimization.}
    \label{fig:method}
\end{figure*}

In this section, we begin by presenting the motivation behind our approach and outlining both the key challenges and the intuitive foundation of our pipeline. We then provide a formal problem definition and describe each component of the framework in detail. Finally, we present a comprehensive illustration of the two training phases, which is shown in \autoref{fig:method}.

\textbf{Stage I: WM and Policy Pretraining.}
In the first stage, we pretrain the world model on offline datasets so that it can capture environment dynamics. In parallel, we pretrain the VLA policy to produce stable action chunks, which serve as a reliable initialization for subsequent optimization.

\textbf{Stage II: VLA Optimization through WM Interaction. }
In the second stage, given an initial frame and a language instruction, the VLA rolls out $n$ action chunks. The world model then interactively generates trajectories conditioned on these actions and provides verified rewards. Using these feedback signals, the VLA is fine-tuned with GRPO to progressively improve policy performance.

\subsection{Problem Formulation}
In this work, we investigate how to train a dual-system VLA policy equipped with a flow-matching action head, using both a WM and a verified reward mechanism. Specifically, we formulate the entire training process as a \textit{Partially Observable Markov Decision Process (POMDP)}. The training pipeline is formally defined by the tuple
\[
\mathcal{M} := (\,\mathcal{O},\,\mathcal{S},\,\mathcal{A},\,\mathcal{L}).
\]
where Observations $\mathcal{O}$ represents the perceptual space of the agent, including real images captured from the environment. States $\mathcal{S}$  denotes the robot’s proprioceptive state. Actions $\mathcal{A}$ is the action space. Language $\mathcal{L}$ refers to natural language instructions provided to the agent.

The VLA policy is expected to generate a sequence of $T$ actions with indices $t \in [T]=\{0,\dots,T\}$, conditioned on the first observed real image $o_i$, the initial language instruction $l_i$, and the initial robot state $s_i$. This process is factorized as
\begin{equation}
\hat{a}_{i:i+T-1} \sim \pi_\theta\!\big(\cdot \mid o_i, l_i, s_i\big)
= \pi_{\theta_{\mathrm{fm}}}\!\big(\cdot \mid z_i, s_i\big), 
\qquad
z_i = f_{\mathrm{VLM}}(o_i, l_i).
\end{equation}
where $f_{\mathrm{VLM}}$ denotes the vision–language large model that encodes multimodal inputs into latent representations $z_i$, and $\pi_{\theta_{\mathrm{fm}}}$ represents the flow-matching policy head that generates the corresponding action chunk.

The world model acts as an interactive simulator that generates rollouts conditioned on the first image $o_t$ and the policy-generated action sequence $a_{t:t+T-1}$. By comparing the generated trajectory against ground-truth images or ground-truth-action-induced rollouts, we obtain a verified reward signal:
\begin{equation}
\hat{o}_{i+t+1} =
\begin{cases}
g_\phi(o_i, a_i), & t=0, \\[6pt]
g_\phi(o_{i:i+t}, a_{i:i+t}), & t=1,\dots,T-1.
\end{cases}
\end{equation}

where $g_\phi$ denotes the autoregressive world model. In particular, the first prediction is generated from the initial frame $o_t$ and the first action $a_t$, while subsequent predictions ($i \geq 1$) are produced autoregressively by conditioning on both the previously generated frames $o_{t:t+i}$ and the executed actions $a_{t:t+i}$.

\subsection{Stage I: WM Pretraining and VLA Pretraining}
To reduce reinforcement learning instability and prevent early collapse, we pretrain the world model and policy on offline datasets, providing a stable initialization for subsequent optimization.

\textbf{World Model Training.}
To obtain dense verified rewards more efficiently, and inspired by recent advances in video generation models (e.g., iVideoGPT~\citep{ivideogpt}), we train an interactive video prediction model to serve as the world model. 
This design avoids the limitations of implicit world models, such as sparse reward signals and the lack of verifiable environment dynamics.
It consists of a pretrained tokenizer and an autoregressive Transformer backbone. During pretraining, the WM is optimized via maximum likelihood:
\begin{equation}
\label{eq:wmloss}
\mathcal{L}^{\text{WM}}_{\text{MLE}}(\phi)
= -\,\mathbb{E}\Big[
\log p_\phi(o_{i+1}\mid o_i, a_i)
+ \sum_{t=1}^{T-1} \log p_\phi(o_{i+t+1}\mid o_{i:i+t}, a_{i:i+t})
\Big].
\end{equation}

where $p_\phi(\cdot)$ denotes the predictive distribution of future observations parameterized by the world model with parameters $\phi$.

\textbf{VLA Pretraining.}
In this stage, we aim to ensure that the VLA produces stable actions. Since the flow-matching action head provides stable training for continuous actions, we pretrain the upstream VLM encoder and the flow-matching head on the expert demonstration dataset $\mathcal{D}$.
\begin{equation}
\label{eq:vla_loss}
\mathcal{L}^{\text{VLA}}_{\text{MSE}}(\theta)
=\;\mathbb{E}_{(a_{i:i+T-1},o_i,l_i,s_i) \sim D}\Big[
\|\,\mathbf{v}_\theta(o_i,l_i,s_i,a^\tau_{i:i+T-1})-u_\tau\|_2^2
\Big].
\end{equation}
where $\tau \sim \mathrm{Beta}(\alpha,\beta)$ is the flow-matching timestep,  
$v_\theta(\cdot)$ denotes the flow predicted by the action head parameterized by $\theta$,  
$a^\tau_{t:t+T-1} = \tau a_{t:t+T-1}+(1-\tau)\epsilon$ is the noise-perturbed action chunk,  
$u_\tau = a_{t:t+T-1}-\epsilon$ is the target flow field defined by the noisy action interpolation,  
and $\epsilon \sim \mathcal{N}(0,I)$ is standard Gaussian noise.

\subsection{Stage II: VLA Optimization through WM Interaction}
To achieve stable and efficient fine-tuning, we adopt an SDE-based policy formulation optimized with GRPO, which offers reliable gradient estimates. The Stage I world model serves as an interactive simulator, providing verified rewards that further enhance training stability.

\textbf{SDE-Policy: Policy Parameterization via Flow and Sigma.}
Since flow matching is inherently a deterministic ODE process, it has limitations in directly obtaining log-likelihood. To address this, we build upon prior work on flow-matching reinforcement learning(e.g. ReinFlow~\citep{reinflow}) by extending the framework into a stochastic formulation, thereby enabling exploration during training. In Stage II, we introduce a \textit{Sigma Net}, whose architecture mirrors that of the flow-matching head, and which outputs a variance vector that parameterizes the stochasticity of the policy. Concretely, at inference time, we discretize the integration into $K=10$ steps, with $k \in {[0,1,2,\dots,10]}$. Actions are generated by integrating the learned vector field from $\tau=0$ to $\tau=1$, initialized from random noise $a_{i:i+T-1}^{\tau=0}\sim \mathcal{N}(0,I)$. We apply the forward Euler method:
\begin{equation}
\mu_k = a^{k\delta }_{i:i+T-1} + \delta  \mathbf{v}_\theta(o_i,l_i,s_i,a^{k\delta}_{i:i+T-1}),
\end{equation}
where $\delta = 0.1$ is the integration step size. For each integration steps $k$, \textit{Sigma Net} takes as input $(z_i, s_i, k)$ and outputs a variance vector $\sigma_\psi^{k}$, while the flow-matching action head simultaneously predicts the flow $\mu_k$. Together, these two components define a Gaussian conditional distribution from which the next action chunk is sampled, thereby generalizing the deterministic FM-ODE formulation into a stochastic differential equation (SDE) process:
\begin{equation}
\label{eq:action}
a^{k\delta}_{i:i+T-1}\sim\mathcal{N}(\mu_k,\Sigma_k),
\end{equation}
where
\begin{equation}
\Sigma_k=(\sigma^{{k}}_\psi)^2.
\end{equation}

Within the same rollout, we compute the step-wise log-likelihoods across the $K$ denoising steps, and take their average as the log-probability of the rollout:
\begin{equation}
\label{eq:log}
\bar{\ell}_{\theta,\psi}=\tfrac{1}{K}\sum_{k=1}^{K}
\log p^{(k)}_{\theta,\psi}(a^{k\delta}_{i:i+T-1}\mid a^{(k-1)\delta}_{i:i+T-1},z_i,s_i).
\end{equation}
Finally, we compute the policy ratio with respect to the old policy by exponentiating the difference of average log-probabilities:
\begin{equation}
r=\exp\!\big(\bar{\ell}_{\theta,\psi}-\bar{\ell}_{\mathrm{old}}\big).
\end{equation}

\begin{algorithm}[t]
\caption{VLA Fine-Tuning Pipeline with World Model and Verified Reward}
\label{alg:vla}
\begin{algorithmic}[1]

    \Require Offline dataset $\mathcal{D}$, diffusion horizon $K$, chunk length $T$, rollout number $N$, initial frame $o_t$, sigma net parameters $\psi$
    \Ensure Trained VLA policy $\pi_{\theta}$

    \State \textbf{Stage I: Pretraining}

    \State Train WM parameters $\phi$ with maximum likelihood Eq.~\ref{eq:wmloss}
    \State Train VLA encoder $f_{\mathrm{VLM}}$ + flow-matching head $\pi_{\theta_{\mathrm{fm}}}$ with loss Eq.~\ref{eq:vla_loss}
    \State \textbf{Stage II: Interaction and Optimization}
    \For{each task instance}
        \For{$n = 1$ to $N$} \Comment{Rollouts}
            \For{$k = 1$ to $K$} \Comment{Diffusion steps}
                \State Sample actions from Gaussian distribution $p^{(k)}_{\theta,\psi}$ \Comment{Eq.~\ref{eq:action}}
                \State Calculate log-probability $\ell^{(k)}$ \Comment{Eq.~\ref{eq:log}}
            \EndFor
            \State Generate trajectory $\mathrm{Traj}$ with WM \Comment{Eq.~\ref{eq:traj_gen}}
            \State Compute verified reward $R_n$ \Comment{Eq.~\ref{eq:reward}}
            \EndFor
        \State Compute advantages $\mathrm{Adv}_n = R_n - \bar{R}_{\mathrm{group}}$
        \State Update policy $\pi_{\theta}$ and sigma net with GRPO objective \Comment{Eq.~\ref{eq:grpo}}
    \EndFor

\end{algorithmic}
\end{algorithm}

\textbf{Interactive WM Simulation and Verified Reward.}
Visual features often carry richer semantic information. To leverage this, given an action chunk $a^{K}_{t:t+T-1}$ from the SDE-Policy, the world model generates a visual trajectory, which is aligned with ground-truth data to construct verified rewards. This design improves reward reliability, reduces manual labeling, and enhances stability.

Starting from the initial frame $o_i$ and the first action $a^{K}_i$, the WM generates the next frame and recursively conditions on previously generated frames to produce the complete trajectory:

\begin{equation}
\label{eq:traj_gen}
\mathrm{Traj}=\big[o_i,\, a^{K\delta}_{i},\, \hat{o}_{i+1},\, \ldots,\, a^{K\delta}_{i+T-1},\, \hat{o}_{i+T}\big], 
\end{equation}
The generated sequence ${\hat{o}_{i+1:i+T+1}}$ is aligned with the ground-truth frames ${o_{i+1:i+T+1}}$ from the offline dataset. The verified reward for the current trajectory segment is defined as the negative weighted sum of the per-frame reconstruction loss and perceptual similarity loss:
\begin{equation}
\label{eq:reward}
R \;=\; - \sum_{t=0}^{T-1}
\Big[\, \lambda_{1}\,L_{1}\!\big(\hat{o}_{i+t+1},\,o_{i+t+1}\big)
\;+\;\lambda_{\mathrm{lp}}\,\operatorname{LPIPS}\!\big(\hat{o}_{i+t+1},\,o_{i+t+1}\big)\Big].
\end{equation}

To reduce variance, we group $n$ rollouts sampled from the same starting state and compute the group average reward as a baseline:

\begin{equation}
\bar{R}_{\mathrm{group}}=\frac{1}{N}\sum_{j=1}^{N} R_j,
\qquad
\mathrm{Adv}_n = R_n - \bar{R}_{\mathrm{group}} .
\end{equation}
Using the policy ratio $r$ derived earlier, the VLA policy is optimized with GRPO. For training stability, we also retain a small-weight flow-matching MSE term as auxiliary supervision on the flow head. The final objective is
\begin{equation}
\label{eq:grpo}
\mathcal{L}^{\mathrm{VLA}}_{\mathrm{GRPO}}(\theta,\psi)
= -\,\mathbb{E}\!\left[\;\operatorname{clip}(r,\,1-\epsilon,\,1+\epsilon)\;\mathrm{Adv}\;\right]
\;+\;\lambda_{\mathrm{mse}}\;\mathcal{L}^{\text{VLA}}_{\text{MSE}}(\theta)
\;-\;\alpha\,\mathbb{H}\!\big(\pi_{\theta,\psi}\big).
\end{equation}
where
$\mathcal{L}^{\text{VLA}}_{\text{MSE}}(\theta)$ is the auxiliary flow-matching MSE loss with weight $\lambda_{\text{mse}}$,  
and $\mathbb{H}(\pi_{\theta,\psi})$ is the policy entropy used to encourage exploration, weighted by $\alpha$. Therefore, the objective integrates policy optimization with auxiliary supervision to ensure efficient and stable fine-tuning.

\section{Experiments}
In this section, we assess \methodname\ through three research questions:
1) How well can world model approximate a simulator?
2) How does world model improve VLA performance?
3) Which components of \methodname\ drive these improvements?

\subsection{Experimental Setup.}

\textbf{Implementations. }
\textbf{1) Benchmark}: We evaluate our model on the LIBERO benchmark \citep{libero}.
\textbf{2) Metrics}: We report success rate (SR) for all tasks.
\textbf{3) Base Policy}: To accelerate experimentation, we employed a lightweight variant of VLA-Adapter \citep{vlaadapter} as our baseline. More details of policy choice can be found in Appendix~\ref{modelarchi}.
\textbf{4) World Model}: To optimize the balance between training efficiency and generation quality, we implemented a lightweight autoregressive world model based on the LLaMA architecture \citep{llama}. This model was instantiated as a compact 138M-parameter variant, comparable in scale to GPT-2 small \citep{gpt2}. The model underwent pretraining on the LIBERO dataset to effectively capture task-relevant visual and action dynamics.
\textbf{5) Training Details}: We initially pretrained a base policy through supervised fine-tuning. Subsequently, we conducted post-training with reinforcement fine-tuning (RFT) using VERL \citep{sheng2025verl}, a distributed RL framework that coordinates diverse rollout strategies with FSDP-sharded training. All experiments were executed on 4× A800 GPUs.

\subsection{World Model capabilities.}

\textbf{Experimental Setting.} To evaluate whether pre-training enables the world model to capture environmental dynamics, we assess its pixel-level generation capability. We randomly sample $T$ consecutive image-action pairs from LIBERO, input the initial frame and complete action sequence into the world model, and compare the generated frames with ground-truth images for subsequent steps.

\textbf{Results Analysis.} As shown in \autoref{tab:wm_performance}, the world model attains low reconstruction error (MSE 0.0039) and strong perceptual scores (PSNR 25.23 dB, SSIM 0.906, LPIPS~\citep{lpips} 0.059), indicating high frame fidelity and perceptual quality. Qualitative results show sharp, temporally consistent frames that capture both static backgrounds and action-driven changes, demonstrating that pre-training enables the model to learn visual appearance and action-conditioned dynamics.

\begin{table}[ht]
    \caption{\textbf{World model generation performance.}
    Left: frame-level metrics across four suites (\emph{Spatial}, \emph{Object}, \emph{Goal}, \emph{Long}) and their averages—MSE (pixel error), PSNR (signal-to-noise ratio), SSIM (structural similarity), and LPIPS (perceptual distance).
    Right: qualitative results. Left column shows simulator sequences, right column shows world-model generations from the same initial frame and actions, illustrating consistent appearance and action-induced dynamics.}
    \label{tab:wm_performance}
    \vspace{-5pt}
    \begin{minipage}{0.5\textwidth}
        \centering
        \setlength{\tabcolsep}{6pt}
        \resizebox{0.90\textwidth}{!}{%
        \begin{tabular}{l|c|c|c|c} 
        \toprule
        \textbf{Task}           & \textbf{MSE $\downarrow$} & \textbf{PSNR $\uparrow$} & \textbf{SSIM $\uparrow$} & \textbf{LPIPS $\downarrow$} \\ 
        \midrule
        Spatial                 & 0.0039 & 24.98 & 0.896 & 0.067  \\ 
        Object                  & 0.0036 & 25.13 & 0.913 & 0.054  \\ 
        Goal                    & 0.0024 & 26.99 & 0.929 & 0.040  \\
        Long                    & 0.0056 & 23.83 & 0.885 & 0.074  \\
        Avg                     & 0.0039 & 25.23 & 0.906 & 0.059  \\
        \bottomrule
        \end{tabular}}
    \end{minipage}
    \hfill
    \begin{minipage}{0.5\textwidth}
        \centering
        \includegraphics[width=1.0\textwidth]{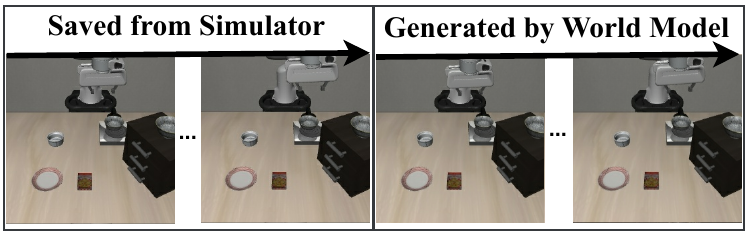} 
    \end{minipage}
\end{table}

\subsection{Performance Improvements for VLA.}
In the previous section, we analyzed the generation quality of the world model.
Here, we further investigate whether our training pipeline enhances policy capability. Specifically, we evaluate policy performance before and after training under the following two task settings.

\begin{wrapfigure}{r}{0.5\textwidth}
    \centering
    \resizebox{0.5\textwidth}{!}{%
    \includegraphics{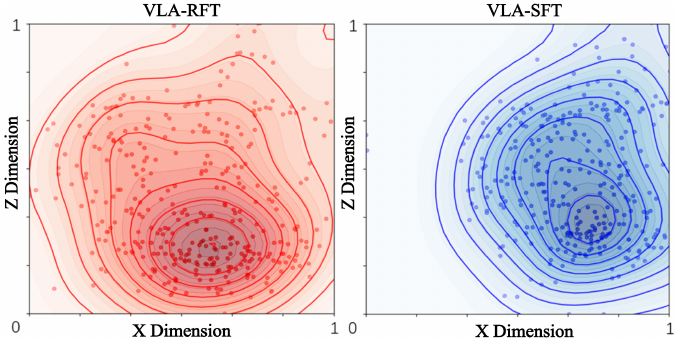}
    }
    \caption{\textbf{Action distribution visualization of VLA-RFT and VLA-SFT.} 
    The plots show distributions along $X$ and $Z$ action dimensions: the left plot corresponds to the RFT-trained policy, and the right plot to the SFT-only base policy.}
     \vspace{-1em}
    \label{fig:ac_dis_shift}
\end{wrapfigure}

\textbf{LIBERO Standard Suites.}
We use the Base ( $15\text{w}$) as baseline and test the effect of adding RFT. As shown in \autoref{tab:contrast_exp}, only $400$ iterations of RFT raise average SR from $86.6\%$ to $91.3\%$ ($+4.7$ points), with gains across all suites: Spatial ($+6.0$ points), Object ($+6.4$ points), Goal ($+2.6$ points), and Long ($+3.0$ points). The graph further shows RFT ($400$) consistently outperforms Base ($15\text{w}$). Notably, while extending SFT from $3\text{w}$ to $15\text{w}$ required heavy training, RFT delivers clear improvements with far fewer iterations, underscoring its efficiency.

\begin{table}[ht]
    \caption{\textbf{Performance under LIBERO Standard Suites.} 
    The table reports SR (\%) across the four suites (\emph{Spatial}, \emph{Object}, \emph{Goal}, and \emph{Long}) and their average; the radar plot on the right provides a visual comparison of different model stages across tasks.}
    \label{tab:contrast_exp}
    \vspace{-5pt}
    \begin{minipage}{0.6\textwidth}
        \centering
        \setlength{\tabcolsep}{6pt}
        \resizebox{1.05\textwidth}{!}{%
        \begin{tabular}{l|c|c|c|c|c} 
        \toprule
        \textbf{Policy (iterations)} & \textbf{Spatial} & \textbf{Object} & \textbf{Goal} & \textbf{Long} & \textbf{Average} \\ 
        \midrule
        Base (3w)       & 82.4 & 84.8 & 85.4 & 57.2 & 77.5 \\ 
        Base (15w)      & 88.4 & 88.0 & 92.8 & 77.2 & 86.6 \\ 
        \methodname (400)        & \textbf{94.4} & \textbf{94.4} & \textbf{95.4} & \textbf{80.2} & \textbf{91.1} \\
        $\Delta$ vs Base (15w)              & \textcolor{ForestGreen}{\textbf{+6.0}} & 
                                \textcolor{ForestGreen}{\textbf{+6.4}} & 
                                \textcolor{ForestGreen}{\textbf{+2.6}} & 
                                \textcolor{ForestGreen}{\textbf{+3.0}} & 
                                \textcolor{ForestGreen}{\textbf{+4.5}} \\
        \bottomrule
        \end{tabular}}
    \end{minipage}
    \hfill
    \begin{minipage}{0.32\textwidth}
        \centering
        \includegraphics[width=0.85\textwidth]{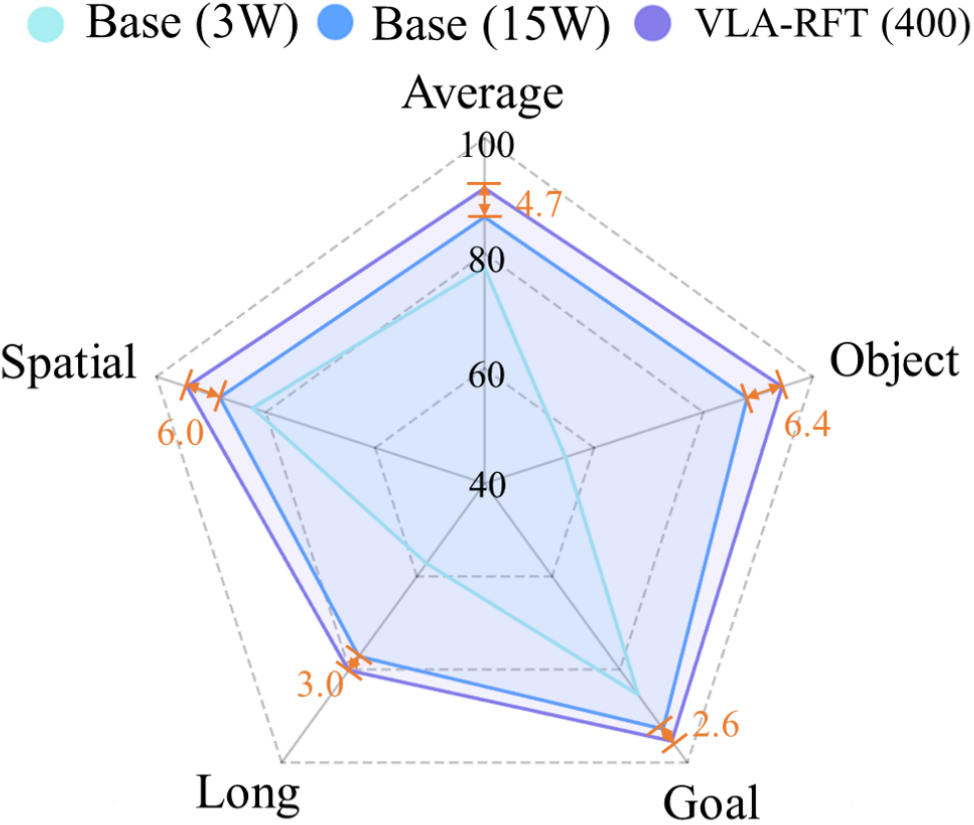} 
    \end{minipage}
\end{table}

\textbf{LIBERO Perturbation Suites.}
To assess out-of-distribution robustness, we construct perturbed variants across the four LIBERO suites and report success rates for base policy and our method. 

\begin{figure}[ht]
    \centering
    \resizebox{\textwidth}{!}{%
    \includegraphics{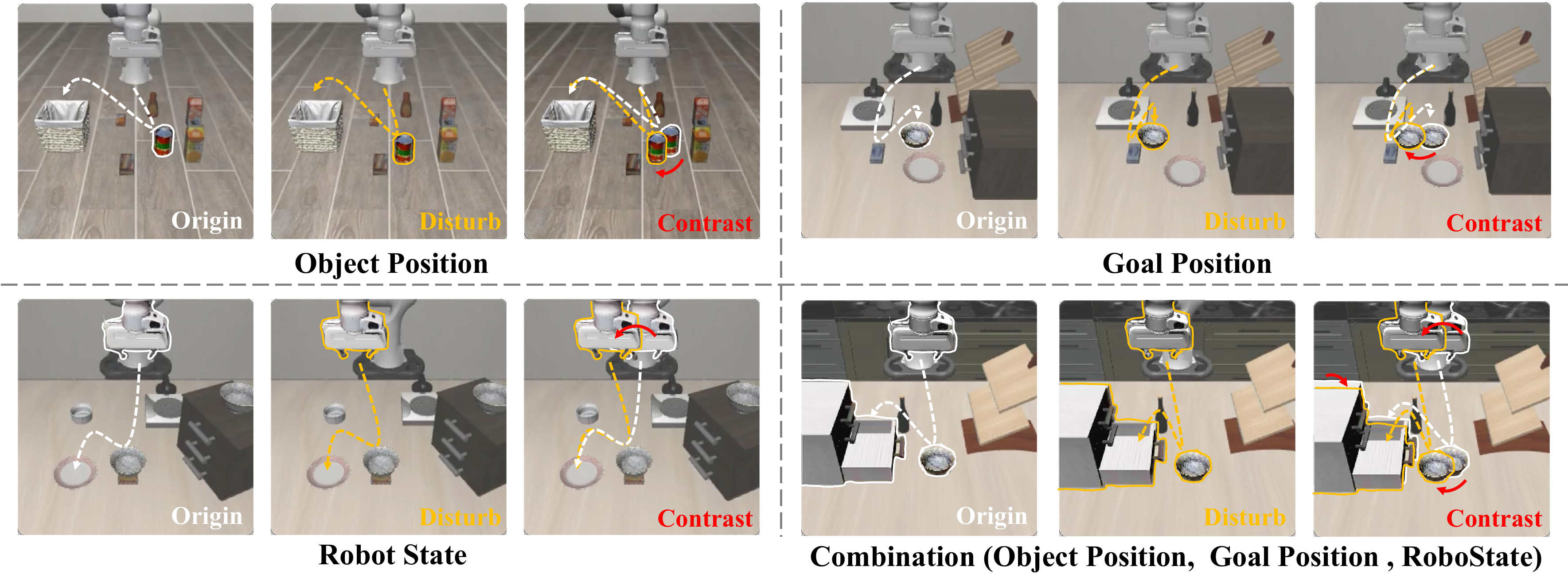}
    }
    \caption{\textbf{Illustration of perturbed task settings in LIBERO.} 
    We consider four perturbation types to evaluate out-of-distribution robustness: 
    (\emph{Object Position}) shifting the initial $(x,y)$ coordinates of the manipulated object; 
    (\emph{Goal Position}) displacing the target object in the $(x,y)$ plane; 
    (\emph{Robot State}) modifying the gripper’s vertical height and horizontal offset; 
    and (\emph{Combination}) applying all perturbations together. 
    Each row shows the original setting (Origin), the perturbed variant (Disturb), and a side-by-side comparison (Contrast).}
    \label{fig:perturbation_setting}
\end{figure}

\textbf{1) Experimental Setting.} In \emph{LIBERO-Object}, the manipulated object’s initial position is shifted in the $(x,y)$ plane with small or large offsets. In \emph{LIBERO-Goal}, the target object’s initial position is similarly displaced. In \emph{LIBERO-Spatial}, the robot’s initial state is perturbed by adjusting the gripper height and horizontal offset. In \emph{LIBERO-Long}, we combine all the above perturbations. An illustration of the perturbed tasks is provided in \autoref{fig:perturbation_setting}.

\textbf{2) Results Analysis.} As shown in \autoref{tab:perturb_result_1}, \methodname consistently improves robustness across all types of perturbations. While the base policy (Base $15\text{w}$) degrades substantially under larger shifts, \methodname maintains higher stability, demonstrating its effectiveness against distributional shifts. The gains are most pronounced in the Goal and combined perturbations (over $+6\%$), where generalization is more challenging, while RoboState perturbations show smaller but consistent improvements. Overall, our training pipeline not only increases standard performance but also improves out-of-distribution robustness, particularly in more complex settings. To further understand the robustness gains, we examine action distributions in \autoref{fig:ac_dis_shift}. \methodname yields broader coverage across action dimensions than base policy, while SFT remains narrowly concentrated. This broader exploration enables better adaptability and generalization under perturbations.

\begin{table}[t]
\centering
\caption{\textbf{Performance under perturbation settings.} All perturbation magnitudes are in centimeter.}
\setlength{\tabcolsep}{6pt}
\renewcommand{\arraystretch}{0.65}
\begin{minipage}[t]{0.48\linewidth}
\centering
\small
\begin{tabularx}{\linewidth}{L|C|C}
\toprule
\textbf{Object Pos Perturb}& \textbf{Range} & \textbf{SR (\%)} \\
\midrule
\rowcolor{gray!15}\multicolumn{3}{c}{\textbf{Minor Perturbation}} \\
Base(15w) & $\pm2.5$ & 69.3 \\
\methodname & $\pm2.5$ & 73.5 \\
$\Delta$ vs Base & $\pm2.5$ &\textcolor{ForestGreen}{\textbf{+4.2}} \\
\midrule
\rowcolor{gray!15}\multicolumn{3}{c}{\textbf{Major Perturbation}} \\
Base(15w) & $\pm5$ & 48.0 \\
\methodname & $\pm5$ & 52.5 \\
$\Delta$ vs Base & $\pm5$ & \textcolor{ForestGreen}{\textbf{+4.5}} \\
\bottomrule
\end{tabularx}
\end{minipage}\hfill
\begin{minipage}[t]{0.48\linewidth}
\centering
\small
\begin{tabularx}{\linewidth}{L|C|C}
\toprule
\textbf{Goal Pos Perturb} & \textbf{Range} & \textbf{SR (\%)} \\
\midrule
\rowcolor{gray!15}\multicolumn{3}{c}{\textbf{Minor Perturbation}} \\
Base(15w) & $\pm2.5$ & 74.5 \\
\methodname & $\pm2.5$ & 79.0 \\
$\Delta$ vs Base & $\pm2.5$ & \textcolor{ForestGreen}{\textbf{+4.5}} \\
\midrule
\rowcolor{gray!15}\multicolumn{3}{c}{\textbf{Major Perturbation}} \\
Base(15w) & $\pm5$ & 44.8 \\
\methodname & $\pm5$ & 51.5 \\
$\Delta$ vs Base & $\pm5$ & \textcolor{ForestGreen}{\textbf{+6.7}} \\
\bottomrule
\end{tabularx}
\end{minipage}

\vspace{1em}

\begin{minipage}[t]{0.48\linewidth}
\centering
\small
\begin{tabularx}{\linewidth}{L|C|C}
\toprule
\textbf{RoboState Perturb}& \textbf{Range} & \textbf{SR (\%)} \\
\midrule
\rowcolor{gray!15}\multicolumn{3}{c}{\textbf{Minor Perturbation}} \\
Base(15w) & $\pm20$ & 73.0 \\
\methodname & $\pm20$ & 76.5 \\
$\Delta$ vs Base & $\pm20$ & \textcolor{ForestGreen}{\textbf{+2.5}} \\
\midrule
\rowcolor{gray!15}\multicolumn{3}{c}{\textbf{Major Perturbation}} \\
Base(15w) & $\pm50$ & 63.5 \\
\methodname & $\pm50$ & 67.0 \\
$\Delta$ vs Base & $\pm50$ & \textcolor{ForestGreen}{\textbf{+3.5}} \\
\bottomrule
\end{tabularx}
\end{minipage}\hfill
\begin{minipage}[t]{0.48\linewidth}
\centering
\small
\begin{tabularx}{\linewidth}{L|C|C}
\toprule
\textbf{Combined Perturb}& \textbf{Range} & \textbf{SR (\%)} \\
\midrule
\rowcolor{gray!15}\multicolumn{3}{c}{\textbf{Minor Perturbation}} \\
Base(15w) & {\scriptsize $\pm2.5/2.5/20$} & 63.5 \\
\methodname & {\scriptsize $\pm2.5/2.5/20$} & 70.0 \\
$\Delta$ vs Base & {\scriptsize $\pm2.5/2.5/20$} & \textcolor{ForestGreen}{\textbf{+6.5}} \\
\midrule
\rowcolor{gray!15}\multicolumn{3}{c}{\textbf{Major Perturbation}} \\
Base(15w) & {\scriptsize $\pm5/5/50$} & 34.0 \\
\methodname & {\scriptsize $\pm5/5/50$} & 37.0 \\
$\Delta$ vs Base & {\scriptsize $\pm5/5/50$} & \textcolor{ForestGreen}{\textbf{+3.0}} \\
\bottomrule
\end{tabularx}
\end{minipage}
\label{tab:perturb_result_1}
\end{table}

\subsection{Key Factors for VLA-RFT}

We showed our pipeline improves policy performance and robustness. Next, we test which components drive these gains via three verified reward designs and world model ablations.

\begin{table}[t]
\centering
\caption{\textbf{Reward design comparison on LIBERO.} 
The left table reports the average success rates (SR, \%) of the base policy (Base 15w) and its variants trained with three different verified reward types. 
The right figure illustrates the corresponding reward function structures.}
\label{tab:reward_exp}

\begin{minipage}{0.44\linewidth}
\centering
\renewcommand{\arraystretch}{1.1}
\resizebox{0.7\linewidth}{!}{%
\begin{tabular}{l|c}
\toprule
\textbf{Policy} & \textbf{Average (SR \%)} \\
\midrule
\rowcolor[HTML]{F2F2F2}\multicolumn{2}{c}{\textbf{Base}} \\
Base (15w) & 86.6 \\
\rowcolor[HTML]{F2F2F2}\multicolumn{2}{c}{\textbf{Reward Type 1}} \\
VLA-RFT (R1) & 87.7 \\
$\Delta$ vs Base & $+1.1$ \\
\rowcolor[HTML]{F2F2F2}\multicolumn{2}{c}{\textbf{Reward Type 2}} \\
VLA-RFT (R2) & 87.1 \\
$\Delta$ vs Base & $+0.5$ \\
\rowcolor[HTML]{F2F2F2}\multicolumn{2}{c}{\textbf{Reward Type 3}} \\
VLA-RFT (Ours) & \textbf{91.1} \\
$\Delta$ vs Base & \textbf{+4.5} \\
\bottomrule
\end{tabular}}
\end{minipage}
\hfill
\begin{minipage}{0.54\linewidth}
\centering
\includegraphics[width=\linewidth]{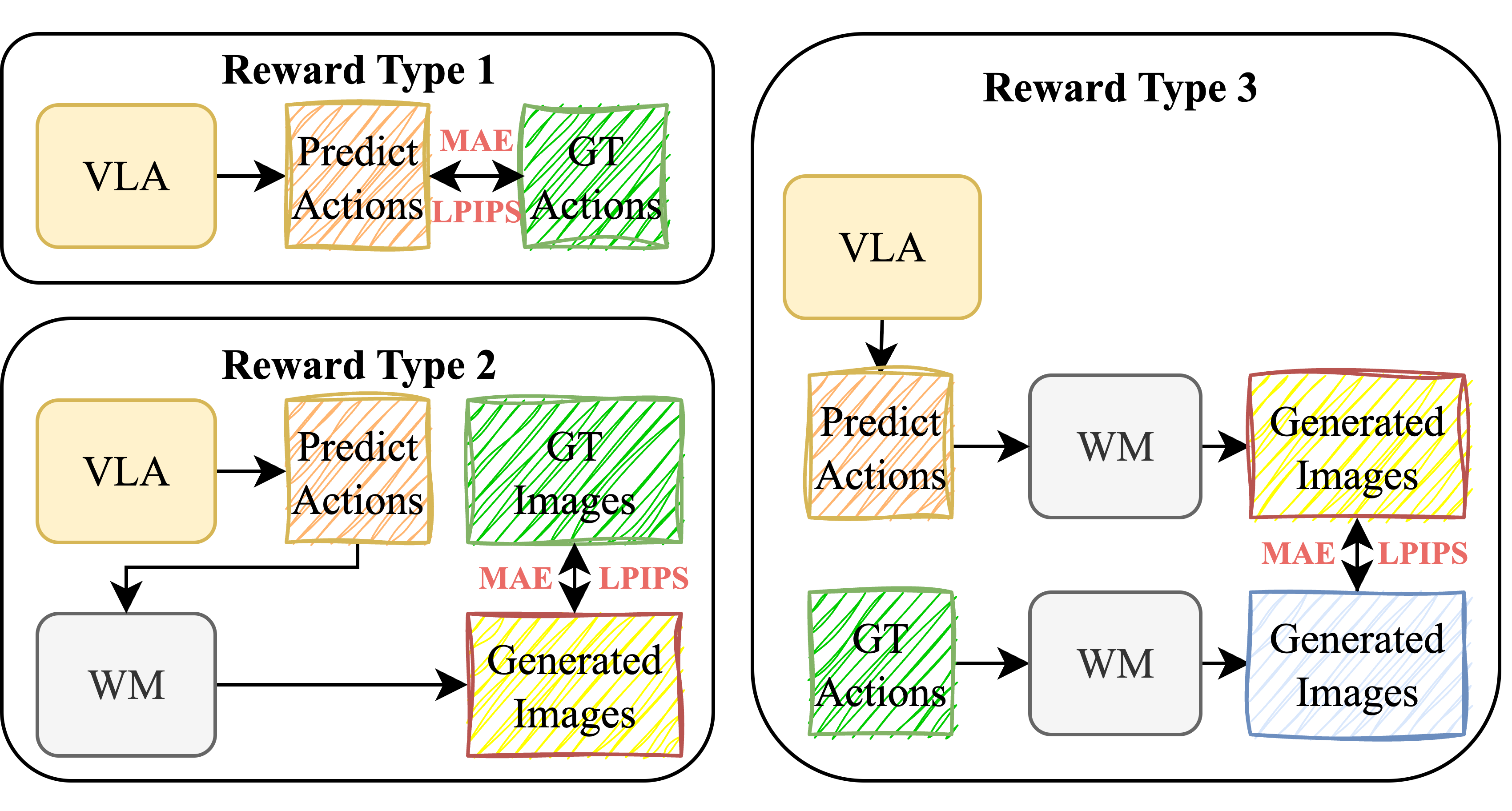} %
\end{minipage}
\end{table}

\textbf{1) Experimental Setting.}
We design three verified rewards under the same training setup and apply RFT to the base model to compare their effects on LIBERO success rates. \emph{\textbf{Reward type 1}} uses the negative L1 distance between policy and dataset actions, offering direct action-level supervision. \emph{\textbf{Reward type 2}} generates images from policy actions via the world model and compares them with dataset images using negative MAE and LPIPS, providing pixel-level guidance. \emph{\textbf{Reward type 3}} renders trajectories from both policy and dataset actions within the same world model, using negative MAE and LPIPS across time to mitigate generation-quality bias and ensure fairness.

\textbf{2) Results Analysis.}
As shown in \autoref{tab:reward_exp}, the comparison across reward designs highlights the essential role of the world model in the training pipeline. \emph{\textbf{Reward type 1}}, which excludes the world model and relies only on action-level supervision, brings very limited gains (+1.1 points), showing that imitation alone is insufficient. \emph{\textbf{Reward type 2}} uses the world model and achieves moderate improvements, but direct comparison with real images still has limitations. \emph{\textbf{Reward type 3}} maximally exploits the world model by performing trajectory comparisons within the same generative space, leading to consistent improvements across all tasks and an average success rate of 91.1\% (+4.5 points over the base policy). These results demonstrate that the world model is a key component, providing reliable optimization signals and enhancing both performance and robustness.

\section{Conclusion \& Limitation}
In this work, we introduced VLA-RFT, a reinforcement fine-tuning framework that uses a learned world model as a controllable simulator. This approach enables efficient and safe policy optimization, bridges imitation and reinforcement learning, and reduces real-world interaction costs. Experiments show strong performance and generalization with minimal fine-tuning, highlighting world-model–based RFT as a promising direction for VLA research.

Despite these advantages, several limitations remain. First, although the WM approximates real-world dynamics and enables reinforcement learning, the verified reward signal is still largely defined by the similarity between generated trajectories and expert demonstrations. Consequently, the policy remains constrained by the quality of the expert dataset, limiting its ability to discover strategies that surpass expert performance. Second, the representational capacity of the WM constitutes a bottleneck; scaling to larger models trained on more diverse and extensive data could significantly improve out-of-distribution generalization. Moreover, our current framework does not explicitly integrate the WM into planning, which could further enhance long-horizon reasoning. Third, the verified reward mechanism itself could be improved: rather than relying solely on expert similarity, future work could leverage learned reward models (e.g., VLAC~\cite{2509.15937}) to provide more task-relevant feedback. Finally, while our study concentrates on flow-matching–based policies, extending the framework to encompass a broader class of policy architectures remains an important direction for future research.


\bibliography{iclr2026_conference}
\bibliographystyle{iclr2026_conference}

\appendix
\clearpage
\section{Appendix}

\subsection{Model Architecture}
\label{modelarchi}

\begin{figure}[ht]
    \centering
    \resizebox{0.8\textwidth}{!}{%
    \includegraphics{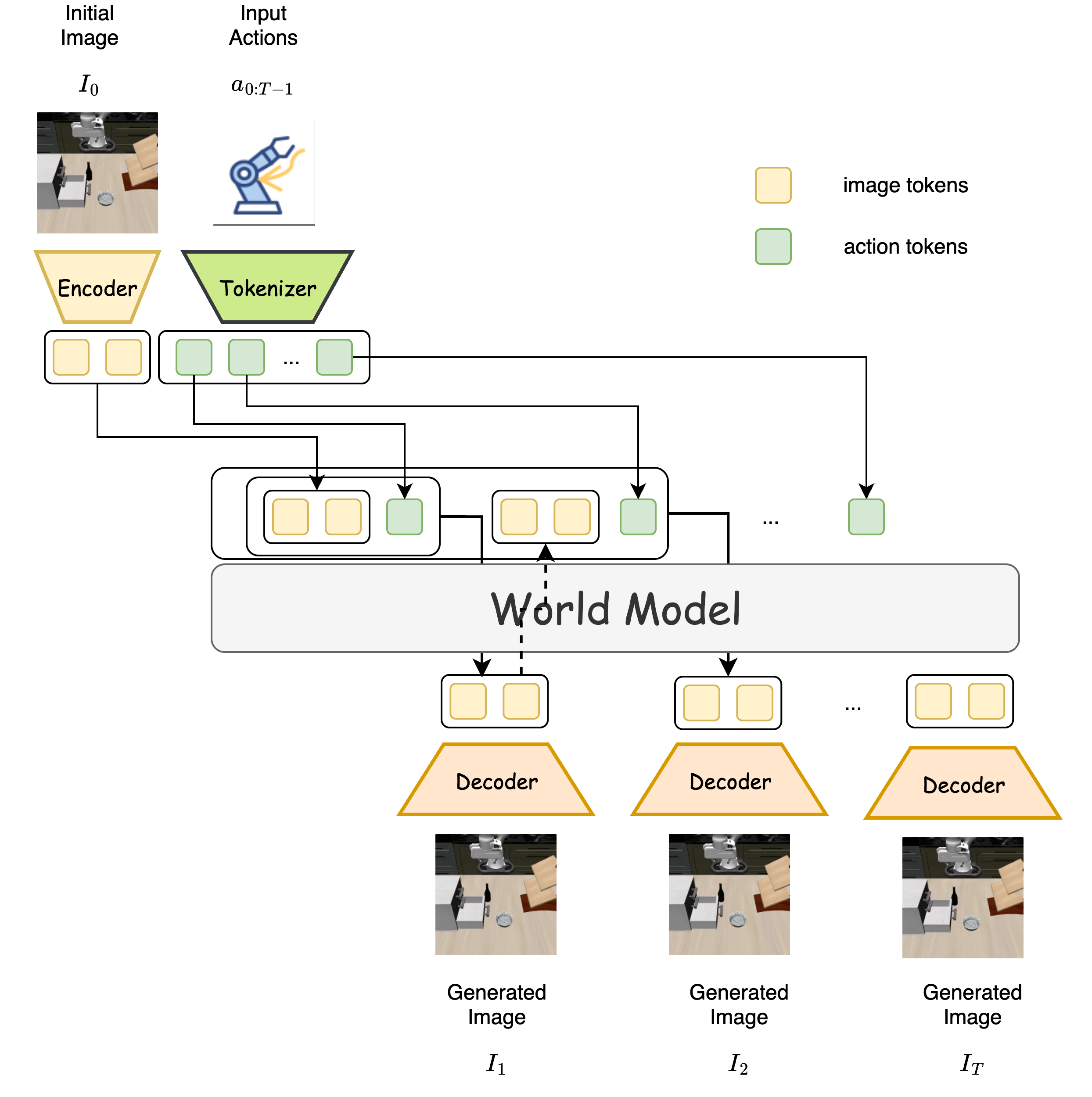}
    }
    \caption{\textbf{Illustration of World Model Generation.} 
    The initial image $I_0$ and input action sequence $a_{0:T-1}$ are first encoded into image and action tokens. These tokens are then fed into the world model to autoregressively predict the future state token sequence. Finally, decoders transform the generated image tokens into predicted future images $I_1, I_2, \dots, I_T$.}
    \label{fig:wm_gen}
\end{figure}

\textbf{World Model.} 
As shown in \autoref{fig:wm_gen}, given the input initial image, we first encode it using an encoder (similar to VQGAN~\citep{vqgen}) to obtain image tokens, while continuous actions are discretized into action tokens through an action tokenizer. These image and action tokens are then jointly fed into the world model, which autoregressively predicts the future token sequences. Finally, the generated image tokens are decoded into corresponding future image sequences, enabling the modeling and simulation of environment dynamics. As shown in \autoref{tab:hparams_two_col}, the model is built on a 12-layer Transformer architecture with a hidden size of 768 and an intermediate FFN size of 3072. It employs 12 attention heads with a head dimension of 64, a maximum positional embedding length of 8192, SiLU activation, and a vocabulary size of 9008. 

\textbf{VLA Policy.} 
While flow-based methods such as $\pi_0$ \citep{pi0} demonstrate competitive performance, their JAX implementation poses integration challenges with VERL, and the LeRobot PyTorch version offers no significant advantages over VLA-Adapter despite its considerable computational overhead. Therefore, we selected VLA-Adapter \citep{vlaadapter} as our base policy. During the RFT stage, we freeze the upper layer VLM of the policy and only update the lower layer action head. In addition, we incorporate a sigma net with a DiT-based architecture similar to the action head, which is responsible for generating noise outputs.

\subsection{Training Details}

\textbf{Pre-Training Phase.} 

1) World Model: As shown in \autoref{tab:hparams_two_col}, the model is optimized using AdamW on the Libero datasets for $1.5 \times 10^5$ steps with a batch size of 16, a segment length of 8, and a learning rate of $5 \times 10^{-5}$. 

2) VLA Policy: Our base policy consists of an upper-layer vision–language model (VLM) and a lower-layer DiT~\citep{dit}-based flow matching action head. During pre-training, we apply LoRA~\citep{lora} for parameter-efficient fine-tuning of the VLM, while jointly optimizing the action head to better align the visual, language, and action spaces. The detailed architecture and training hyperparameters are summarized in \autoref{tab:vla_adapter_hparams}.

\textbf{RFT Phase.} 

\begin{figure}[ht]
    \centering
    \includegraphics[width=\textwidth]{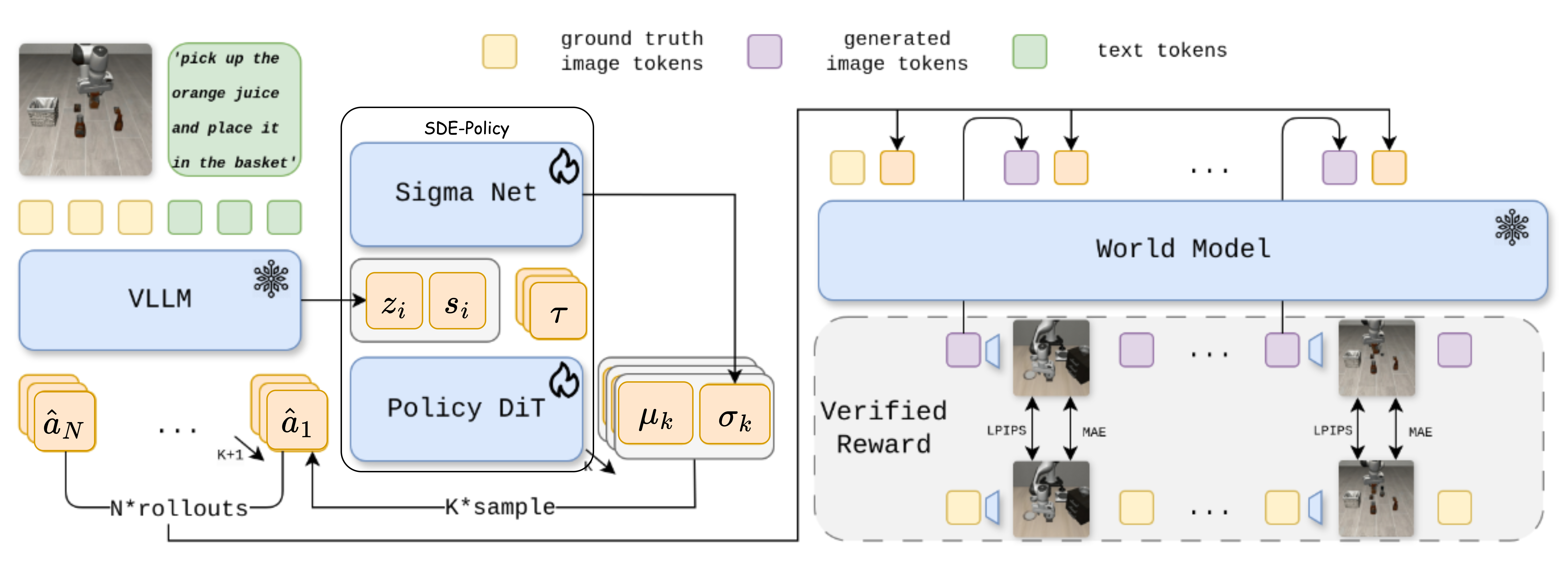}
    \caption{\textbf{Detailed Implementation of Method.} 
    }
    \label{fig:method_apendix}
\end{figure}

For more details, see \autoref{fig:method_apendix}.

1) World Model: The World Model is frozen.

2) VLA Policy: As shown in \autoref{tab:rl_hparams}, we adopt GRPO~\citep{chen2025tgrpo} as the advantage estimator and configure the optimization with a learning rate of $1\times10^{-6}$ and a sigma learning rate of $1\times10^{-5}$. 
For stability, an auxiliary MSE loss is included with coefficient $0.01$, together with an entropy regularization term of $0.003$ to encourage exploration. 
Training is conducted for $400$ steps with a batch size of $16$, and each update uses $16$ rollouts. 
These settings strike a balance between stability and efficiency, enabling consistent improvements under limited compute budgets.

\begin{table}[t]
\centering
\caption{Key hyperparameters of the World Model: \emph{Architecture} (left) and \emph{Pre-training} (right).}
\label{tab:hparams_two_col}
\setlength{\tabcolsep}{6pt}

\begin{minipage}{0.48\linewidth}
\centering
\renewcommand{\arraystretch}{1.1}
\begin{tabular}{@{}l l@{}}
\toprule
\textbf{Hyperparameter} & \textbf{Value} \\
\midrule
\multicolumn{2}{@{}l}{\textbf{Architecture}} \\
\midrule
Layers & 12 \\
Hidden size & 768 \\
FFN intermediate size & 3072 \\
Attention heads & 12 \\
Head dimension & 64 \\
Key--value heads & 12 \\
Max position embeddings & 8192 \\
Activation & SiLU \\
Vocabulary size & 9008 \\
\bottomrule
\end{tabular}
\end{minipage}
\hfill
\begin{minipage}{0.48\linewidth}
\centering
\renewcommand{\arraystretch}{1.1}
\begin{tabular}{@{}l l@{}}
\toprule
\textbf{Hyperparameter} & \textbf{Value} \\
\midrule
\multicolumn{2}{@{}l}{\textbf{Pre-training}} \\
\midrule
Batch size & 16\\
Training steps &$1.5 \times 10^5$ \\
Learning rate & $5 \times 10^{-5}$\\
Optimizer & AdamW ~\citep{adam}\\
Datasets & Libero Datasets\\
Segment length & 8\\
\bottomrule
\end{tabular}
\end{minipage}
\end{table}

\begin{table}[t]
\centering
\caption{Key hyperparameters of the VLA-Adapter: \emph{Architecture} (left) and \emph{Pre-training} (right).}
\label{tab:vla_adapter_hparams}
\setlength{\tabcolsep}{6pt}

\begin{minipage}{0.48\linewidth}
\centering
\renewcommand{\arraystretch}{1.1}
\resizebox{\linewidth}{!}{%
\begin{tabular}{@{}l l@{}}
\toprule
\textbf{Hyperparameter} & \textbf{Value} \\
\midrule
\multicolumn{2}{@{}l}{\textbf{Architecture}} \\
\midrule
Vision backbone & dinosiglip-vit-so-224px \\
Input image size & $224 \times 224$ \\
LLM backbone  & qwen25-0\_5b-extra \\
LLM max length & 2048 \\
Text layers / hidden size & 24 / 896 \\
Attention heads / KV heads & 14 / 2 \\
FFN intermediate size & 4864 \\
Max position embeddings & 32768 \\
Torch dtype & bfloat16 \\
Action bins  & 256 \\
\bottomrule
\end{tabular}}
\end{minipage}
\hfill
\begin{minipage}{0.48\linewidth}
\centering
\renewcommand{\arraystretch}{1.1}
\resizebox{\linewidth}{!}{%
\begin{tabular}{@{}l l@{}}
\toprule
\textbf{Hyperparameter} & \textbf{Value} \\
\midrule
\multicolumn{2}{@{}l}{\textbf{Pre-training}} \\
\midrule
Batch size & 16 \\
Training steps & $1.5\times 10^{5}$ \\
Learning rate & $1\times 10^{-4}$ \\
Optimizer & AdamW~\citep{adam} \\
Datasets & Libero Datasets \\
LoRA Rank & 64 \\
\bottomrule
\end{tabular}}
\end{minipage}
\end{table}

\begin{table}[t]
\centering
\caption{Key hyperparameters for RL fine-tuning.}
\label{tab:rl_hparams}
\renewcommand{\arraystretch}{1.1}
\setlength{\tabcolsep}{6pt}
\begin{tabular}{l l}
\toprule
\textbf{Hyperparameter} & \textbf{Value} \\
\midrule
Advantage estimator & GRPO \\
Learning rate & $1 \times 10^{-6}$ \\
Sigma learning rate & $1 \times 10^{-5}$ \\
MSE loss coefficient & 0.01 \\
Entropy coefficient & 0.003 \\
Total training steps & 400 \\
Batch Size & 16 \\
Rollout Times & 16 \\
\bottomrule
\end{tabular}
\end{table}

\subsection{Experiment Details}

\textbf{Details of perturbation experiments.}  
The details of the perturbation experiments are shown in \autoref{tab:appendix_1}. Task 1 and Task 2 denote different tasks, while Dim 1 and Dim 2 refer to different perturbation objects or robot states.

\begin{table}[t]
\caption{\textbf{Details of perturbation experiments.}Task 1 and Task 2 denote different tasks, while Dim 1 and Dim 2 refer to different perturbation
objects or robot states. Where KP means keep original states.}
\label{tab:appendix_1}
\begin{center}
\resizebox{\textwidth}{!}{
\begin{tabular}{cccc|ccccc}
\toprule
  \begin{tabular}[c]{@{}c@{}}Policy.\end{tabular} &
  \cellcolor{yellow!15}\begin{tabular}[c]{@{}c@{}}Object\\ Position\end{tabular} &
  \cellcolor{green!15}\begin{tabular}[c]{@{}c@{}}Goal\\ Position\end{tabular} &
  \cellcolor{blue!15}\begin{tabular}[c]{@{}c@{}}Robot\\ Initial States\end{tabular} &
  \begin{tabular}[c]{@{}c@{}}Task1 Dim1\\ SR (\%)\end{tabular} &
  \begin{tabular}[c]{@{}c@{}}Task1 Dim2\\ SR (\%)\end{tabular} &
  \begin{tabular}[c]{@{}c@{}}Task2 Dim1\\ SR (\%)\end{tabular} &
  \begin{tabular}[c]{@{}c@{}}Task2 Dim2\\ SR (\%)\end{tabular} &
  \begin{tabular}[c]{@{}c@{}}Average\\ SR (\%)\end{tabular} \\ \midrule
  Base & \cellcolor{yellow!15}$\pm2.5$ & KP& KP& 87& 52& 78& 60& 69.3 \\
  Ours & \cellcolor{yellow!15}$\pm2.5$ & KP& KP& 94& 62& 80& 58& 73.5 \\
    \midrule
  Base & \cellcolor{yellow!15}$\pm5$ & KP& KP& 70& 44& 50& 28& 48.0 \\
  Ours & \cellcolor{yellow!15}$\pm5$ & KP& KP& 72& 52& 56& 30& 52.5 \\
  \midrule
  Base & KP& \cellcolor{green!15}$\pm2.5$& KP& 62& 58& 92& 86& 74.5\\
  Ours & KP& \cellcolor{green!15}$\pm2.5$ & KP& 64& 68& 94& 90& 79.0\\
    \midrule
  Base & KP& \cellcolor{green!15}$\pm5$  & KP& 34& 46& 48& 54& 44.8\\
  Ours & KP& \cellcolor{green!15}$\pm5$  & KP& 46& 42& 58& 60& 51.5\\
  \midrule
  Base & KP& KP& \cellcolor{blue!15} $\pm20$& 60& 88& 54& 90& 73.0\\
  Ours & KP& KP& \cellcolor{blue!15} $\pm20$& 62& 92& 58& 94& 76.5\\
    \midrule
  Base & KP& KP& \cellcolor{blue!15} $\pm50$& 42& 82& 52& 78& 63.5\\
  Ours & KP& KP& \cellcolor{blue!15} $\pm50$& 46& 86& 56& 80& 67.0\\
  \midrule
  Base & \cellcolor{yellow!15}$\pm2.5$ & \cellcolor{green!15}$\pm2.5$ & \cellcolor{blue!15}$\pm20$ & 64& 82& 36& 72& 63.5\\
  Ours & \cellcolor{yellow!15}$\pm2.5$ & \cellcolor{green!15}$\pm2.5$ & \cellcolor{blue!15}$\pm20$ & 68& 92& 40& 80& 70.0\\
  Base & \cellcolor{yellow!15}$\pm5$ & \cellcolor{green!15}$\pm5$ & \cellcolor{blue!15}$\pm50$ & 34& 64& 8& 30& 34.0\\
  Ours & \cellcolor{yellow!15}$\pm5$ & \cellcolor{green!15}$\pm5$ & \cellcolor{blue!15}$\pm50$& 36& 60& 12& 40& 37.0\\
  \bottomrule
\end{tabular}
}
\end{center}
\end{table}

\textbf{Comparisions with other VLA methods.}
As shown in \autoref{tab:sota_exp}, VLA-RFT (Ours) consistently achieves the highest scores compared with baseline policies.

\begin{table}[ht]
    \caption{\textbf{Performance under general settings of LIBERO suites.} 
    We report SR (\%) across the four suites (\emph{Spatial}, \emph{Object}, \emph{Goal}, and \emph{Long}) and their average. VLA-RFT (ours) consistently achieves the highest scores compared with baseline policies. VLA-Adapter (Base) is the recurrence result when the Policy is Flow-matching and there is only one image input.}
    \label{tab:sota_exp}
    \centering
    \resizebox{\textwidth}{!}{%
        \begin{tabular}{l|cc|cc|cc|cc|cc} 
        \toprule
        \multirow{2}{*}{\textbf{Policy}} & 
        \multicolumn{2}{c|}{\textbf{Spatial}} & 
        \multicolumn{2}{c|}{\textbf{Object}} & 
        \multicolumn{2}{c|}{\textbf{Goal}} & 
        \multicolumn{2}{c|}{\textbf{Long}} & 
        \multicolumn{2}{c}{\textbf{Average}} \\ 
        \cmidrule(lr){2-3} \cmidrule(lr){4-5} \cmidrule(lr){6-7} \cmidrule(lr){8-9} \cmidrule(lr){10-11}
        & \textbf{SR (\%)} & \textbf{Rank} 
        & \textbf{SR (\%)} & \textbf{Rank} 
        & \textbf{SR (\%)} & \textbf{Rank} 
        & \textbf{SR (\%)} & \textbf{Rank} 
        & \textbf{SR (\%)} & \textbf{Rank} \\ 
        \midrule
        Diffusion Policy~\citep{diffusion-policy}    & 78.3 & 11 & 92.5 & 4 & 68.3 & 11 & 50.5 & 11 & 72.4 & 11 \\ 
        Octo~\citep{octo}                & 78.9 &  9& 85.7 & 10 & 84.6 & 5 & 51.1 & 10 & 75.1 & 9 \\ 
        MDT~\citep{mdt}                & 78.5 &10  & 87.5 & 9 & 73.5 & 10 & 64.8 & 5 & 76.1 & 8\\ 
        OpenVLA~\citep{openvla}            & 84.7 & 7 & 88.4 & 7 & 79.2 & 7 & 53.7 & 9 & 76.5 & 7\\        
        SpatialVLA~\citep{qu2025spatialvla}  &88.2 &  4  &89.9 &6&78.6 &8&55.5 &7&78.1&6\\
        WorldVLA~\citep{cen2025worldvla} &87.6 &  5  &96.2&2& 83.4&6& 60.0&6 &81.8&4\\
        CoT-VLA~\citep{cot-vla} &87.5& 6  & 91.6&5& 87.6 &4&69.0 &4&81.1&5\\
        TraceVLA~\citep{tracevla}  & 84.6& 8  &85.2 &11& 75.1&9& 54.1&8& 74.8&10\\
        $\pi_0$      ~\citep{pi0}       & \emph{91.2} & 2 & \emph{93.2} & 3 & \emph{93.8} & 2 & 74.2 & 3 & \emph{88.1} & 2 \\
        VLA-Adapter~\citep{vlaadapter} (Base)       & 88.4 & 3 & 88.0 & 8 & 92.8 & 3 & \emph{77.2} & 2 & 86.6 & 3 \\ 
        VLA-RFT (Ours)           & \textbf{94.4} & 1 & \textbf{94.4} & 1 & \textbf{95.4} & 1 & \textbf{80.2} & 1 & \textbf{91.1} & 1 \\
        \bottomrule
        \end{tabular}%
    } 
\end{table}

\textbf{Comparisons with other VLA+RL methods.}
Our comprehensive evaluation demonstrates that the proposed framework achieves remarkable superiority over existing approaches across multiple dimensions. Not only does our method significantly outperform state-of-the-art offline RL baselines, but it also rivals the performance of online RL methods while maintaining the practical advantages of offline training. Most notably, our world-model-based approach delivers these superior results with dramatically reduced computational overhead, requiring substantially fewer training steps than conventional alternatives.
The experimental comparison reveals the distinct advantages of our approach across diverse settings. While VLA-RL operates through direct reinforcement learning in the LIBERO environment, and competing methods like ARFM, RWR, and ReinboT represent the current best practices in offline RL, our framework consistently demonstrates superior performance gains. The key innovation lies in how VLA-RFT strategically exploits the world model's predictive capabilities to achieve unprecedented data efficiency, enabling faster convergence without sacrificing performance quality.
For transparency and reproducibility, we note that VLA-RL results are sourced directly from the original publication, while the performance metrics for ARFM, RWR, and ReinboT on LIBERO are derived from the ARFM paper, ensuring fair and comprehensive benchmarking across all methods.

\begin{table}[h]
\centering
\caption{\textbf{Comparison with other RL methods on Libero Average.} 
We report baseline success rate (SR), fine-tuned SR, their improvement ($\Delta$), and training steps.}
\label{tab:success_rate}
\resizebox{0.8\textwidth}{!}{%
\begin{tabular}{l l c c c c}
\toprule
\textbf{Type} & \textbf{Algorithm} & \textbf{Baseline SR (\%)} & \textbf{SR (\%)} & \textbf{$\Delta$ SR (\%)} & \textbf{Training Steps} \\
\midrule
Online  & VLA-RL~\citep{lu2025vla}         & 76.5 & 81.0 & \textbf{4.5} & 10,000 \\
\midrule
\multirow{3}{*}{Offline} 
        & ARFM~\citep{arfm}            & \textbf{88.1} & \textbf{92.1} & 4.0 & 40,000 \\
        & RWR~\citep{rwr}            & \textbf{88.1} & 90.8 & 2.7 & 40,000 \\
        & ReinboT~\citep{reinbot}         & \textbf{88.1} & 91.2 & 3.1 & 40,000 \\
\midrule
Ours    & \textbf{VLA-RFT} & 86.6 & 91.1 & \textbf{4.5} & \textbf{400} \\
\bottomrule
\end{tabular}}
\end{table}

\begin{figure}[ht]
    \centering
    \resizebox{\textwidth}{!}{%
    \includegraphics{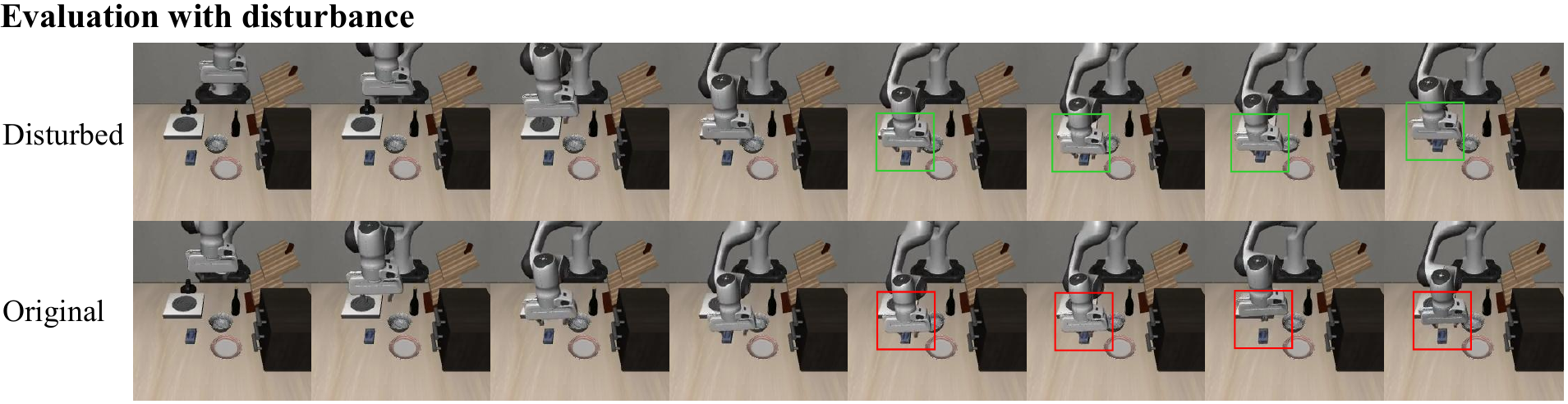}
    }
    \caption{\textbf{Comparison of original and disturbed scenarios.} }
    \label{fig:appendix_rollout_1}
\end{figure}

\begin{figure}[ht]
    \centering
    \resizebox{\textwidth}{!}{%
    \includegraphics{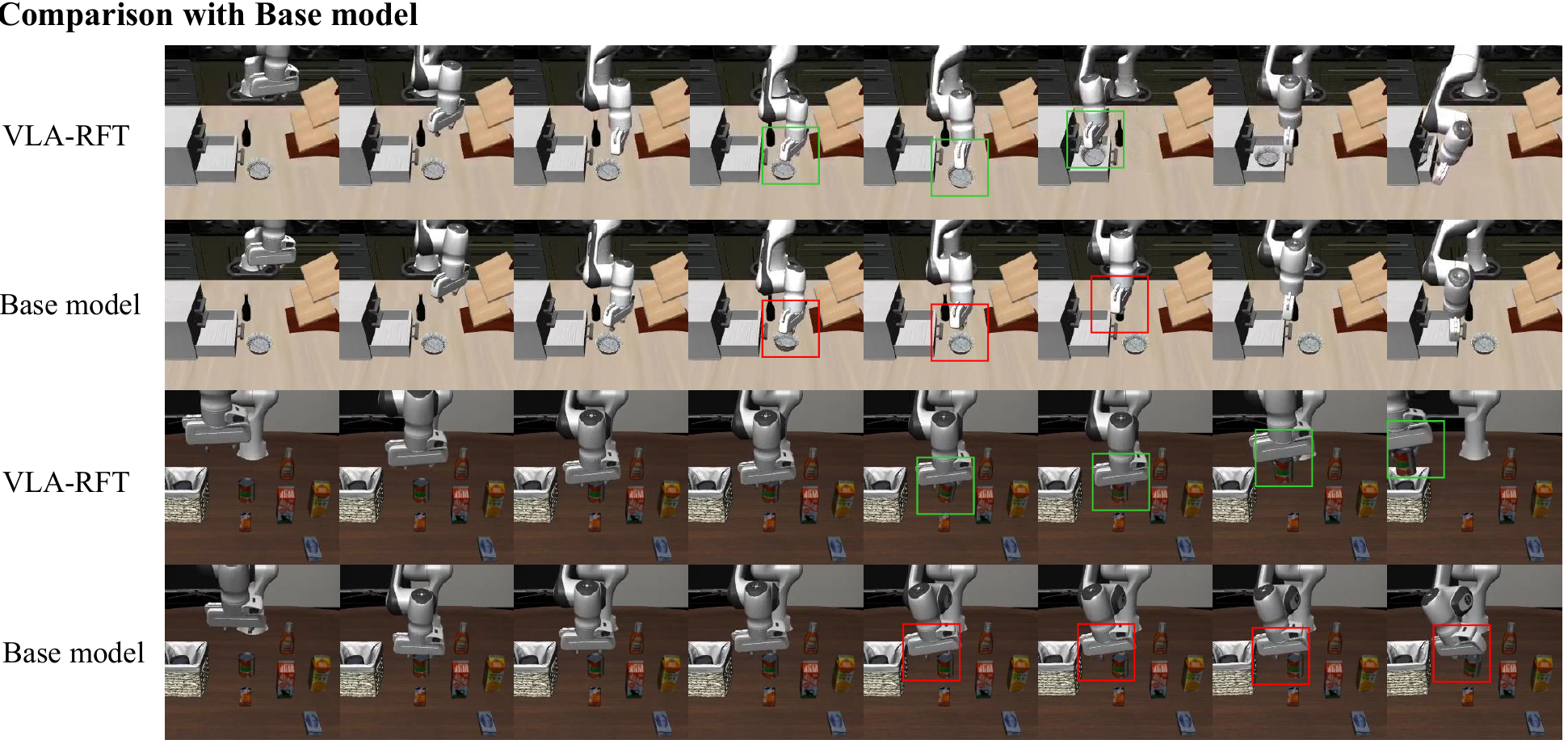}
    }
    \caption{\textbf{Comparison of base policy and VLA-RFT.} }
    \label{fig:appendix_rollout_2}
\end{figure}

\textbf{Visualization.} We also provide more detailed visualization results in \autoref{fig:appendix_rollout_1} and \autoref{fig:appendix_rollout_2}.

\subsection{The Use of Large Language Models (LLMs)}
To enhance the readability and coherence of this paper, we employed large language models to assist in refining the writing.

\end{document}